%% file: main.tex
\documentclass[runningheads]{llncs}
\usepackage[T1]{fontenc}
\input{preamble}
\usepackage{cite}
\usepackage[pagebackref,breaklinks,colorlinks]{hyperref}
\hypersetup{colorlinks, citecolor=blue}
\usepackage{tikz}
\usepackage[capitalize]{cleveref}
\crefname{section}{Sec.}{Secs.}
\Crefname{section}{Section}{Sections}
\Crefname{table}{Table}{Tables}
\crefname{table}{Tab.}{Tabs.}
\usepackage{graphicx}
\usepackage{wrapfig}
\usepackage{mdframed}

\usepackage{gensymb}
\usepackage{listings} 
\usepackage{minted}         
\usepackage{float}          

\usepackage{xcolor}         
\definecolor{LightGray}{rgb}{0.9,0.9,0.9}  
\usepackage{colortbl}  

\usepackage{stackengine}

\usepackage{float}
\usepackage{orcidlink}

\newcolumntype{M}[1]{>{\centering\arraybackslash}m{#1}} 

\newcommand{\repeatthanks}{\textsuperscript{\thefootnote}}
\usepackage[width=122mm,left=12mm,paperwidth=146mm,height=193mm,top=12mm,paperheight=217mm]{geometry}

\begin{document}
\title{RefChartQA: Grounding Visual Answer on Chart Images through Instruction Tuning}
\titlerunning{RefChartQA}

\author{
Alexander Vogel\,\orcidlink{0009-0006-4357-8525}\inst{1,\thanks{Equal contribution. \text{\textdagger} Corresponding author.}} \and
Omar Moured\,\orcidlink{0000-0003-4227-8417}\inst{1,\repeatthanks} \and
Yufan Chen\,\orcidlink{0009-0008-3670-4567}\inst{1} \and
Jiaming Zhang\,\orcidlink{0000-0003-3471-328X}\inst{1,2,\text{\textdagger}} \and \\
Rainer Stiefelhagen\,\orcidlink{0000-0001-8046-4945}\inst{1}
}

\authorrunning{A. Vogel \& O. Moured et al.}

\institute{CV:HCI lab, Karlsruhe Institute of Technology, Germany. \and
CVG, ETH, Switzerland. \\ 
}

\maketitle  
\begin{figure}
    \centering
    \includegraphics[width=1\linewidth]{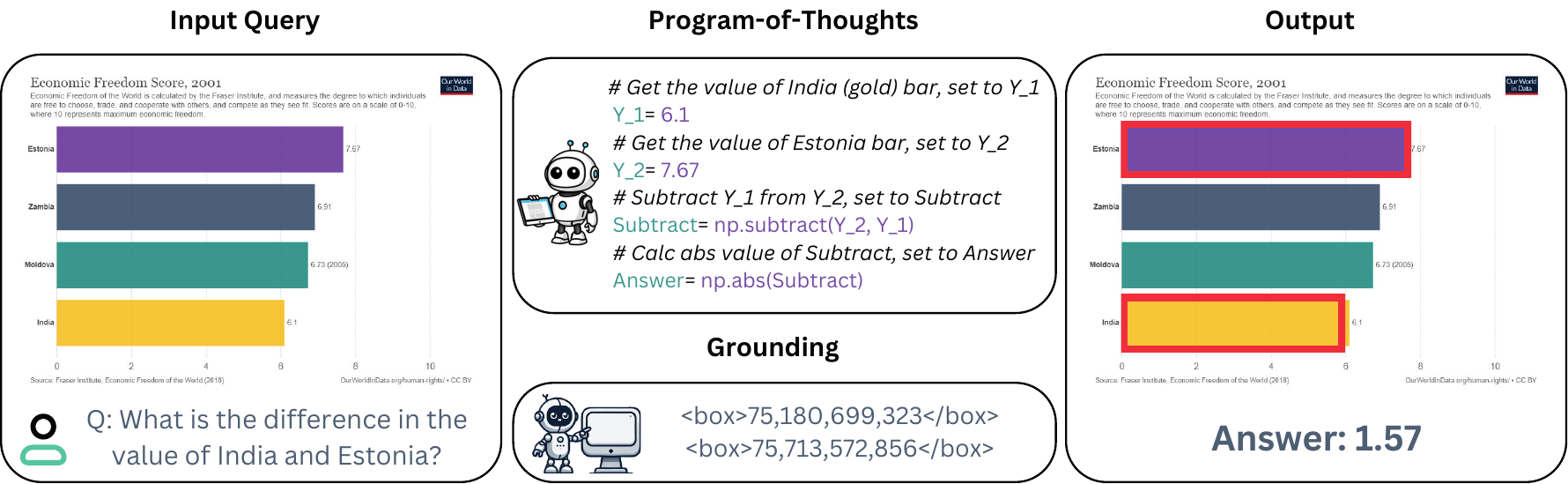}
    \caption{RefChartQA establishes a comprehensive instruction-tuning benchmark designed to integrate chart understanding tasks such as PoT with evidence visual grounding. By introducing spatial text alignment, we demonstrate its role in enhancing the robustness and reliability of MLLMs.
    }
    \vskip -8ex
    \label{fig:refchartqa_overview}
\end{figure}

\input{chapters/abstract}

\input{chapters/introduction}

\input{chapters/relatedwork}

\input{chapters/benchmark}

\input{chapters/method}

\input{chapters/experiment}

\input{chapters/conclusion_and_future_work}

\newpage
\bibliographystyle{splncs04}
\bibliography{main}

\end{document}

%% file: preamble.tex
\usepackage[dvipsnames, table]{xcolor}

\usepackage{diagbox}
\usepackage{array}
\usepackage{mathrsfs}
\usepackage{makecell}
\usepackage{pifont}
\usepackage{placeins}
\usepackage[linesnumbered,ruled,vlined]{algorithm2e}

\usepackage{graphicx}
\usepackage{amsmath}
\usepackage{amssymb}
\usepackage{booktabs}
\usepackage{mathtools}
\usepackage[accsupp]{axessibility}  
\usepackage{color}
\usepackage{array}
\usepackage{pdflscape}
\usepackage[longtable]{multirow}
\usepackage{longtable}
\usepackage{tabularray}
\usepackage{booktabs}
\usepackage{times}
\usepackage{epsfig}
\usepackage{amsmath}
\usepackage{bbm}
\DeclareMathAlphabet\mathbfcal{OMS}{cmsy}{b}{n}
\usepackage{makecell}
\usepackage{siunitx}
\usepackage{svg}
\usepackage{float}
\usepackage{comment}
\usepackage{lipsum}
\usepackage{stfloats}
\usepackage{multirow}
\usepackage{bm}
\usepackage{etoolbox}
\usepackage{icomma}
\usepackage{array}
\usepackage{tabulary}
\usepackage{xcolor}
\usepackage{paralist}
\usepackage{booktabs}
\usepackage{adjustbox}
\usepackage{caption}
\usepackage{arydshln}
\usepackage{rotating}

\definecolor{gray}{rgb}{0.3,0.3,0.3}
\definecolor{blue}{rgb}{0,0.5,1}
\definecolor{mask_red}{rgb}{1,0,0.8}
\definecolor{green}{rgb}{0.2,1,0.2}
\definecolor{rblue}{rgb}{0,0,1}
\definecolor{lightblue}{HTML}{6495ed}
\definecolor{lightred}{HTML}{F19C99}

\definecolor{graytablerow}{gray}{0.6}

\usepackage{pifont}
\usepackage{tabu}
\usepackage[pro]{fontawesome5}

\usepackage{tikz}
\newcommand*\circled[1]{\tikz[baseline=(char.base)]{
\node[shape=circle,fill=gray,inner sep=0.5pt] (char) {\textcolor{white}{\small \textbf{#1}}};}}

\usepackage[frozencache,cachedir=.]{minted}

%% file: chapters/abstract.tex
\begin{abstract}
Recently, Vision Language Models (VLMs) have increasingly emphasized document visual grounding to achieve better human-computer interaction, accessibility, and detailed understanding. However, its application to visualizations such as charts remains under-explored due to the inherent complexity of interleaved visual-numerical relationships in chart images. Existing chart understanding methods primarily focus on answering questions without explicitly identifying the visual elements that support their predictions. To bridge this gap, we introduce RefChartQA, a novel benchmark that integrates Chart Question Answering (ChartQA) with visual grounding, enabling models to refer elements at multiple granularities within chart images. Furthermore, we conduct a comprehensive evaluation by instruction-tuning 6 state-of-the-art VLMs across different categories. Our experiments demonstrate that incorporating spatial awareness via grounding improves response accuracy by over 15\%, reducing hallucinations, and improving model reliability. Additionally, we identify key factors influencing text-spatial alignment, such as architectural improvements in TinyChart, which leverages a token-merging module for enhanced feature fusion. Our dataset is open-sourced for community development and further advancements. All models and code will be publicly available at \url{https://github.com/moured/RefChartQA}.

\end{abstract}

%% file: chapters/introduction.tex
\newpage
\section{Introduction}
\label{sec:intro}

Charts serve as an essential medium for conveying information in numerous documents, including scientific articles, business reports, and educational materials. Recent advances in \textit{chart understanding} have led to promising progress in tasks such as chart visual question answering (Chart VQA) and chart-based information extraction~\cite{masry-etal-2022-chartqa, kantharaj-etal-2022-chart,liu2022matcha}. However, most existing works mainly concentrate on predicting the final answers based on charts, largely overlooking the \emph{interpretability} aspect of chart understanding tasks. To address this, recent work on Visual Answer Grounding (VAG) 
extends VQA by highlighting the specific regions within an image that support the predicted answer\cite{chen2022grounding, zhu2016visual7w}. Although VAG has been explored in natural images, its application to data visualizations remains under-studied. In charts, grounding entails highlighting specific bars, points, or legend entries that justify a numerical or factual response \cite{masry-etal-2022-chartqa}.

In this paper, we propose a novel framework RefChartQA that integrates Chart VQA with VAG, aiming to provide more transparent and credible answers for chart-related queries. Fig.~\ref{fig:refchartqa_overview} shows the overview of our proposed RefChartQA. Specifically, we address the challenges of \circled{1} reliable extraction of textual and graphical elements from diverse chart formats, \circled{2} interpretable reasoning over both numeric and textual data, and \circled{3} grounding answers by visually localizing chart elements that contribute to the predicted response. Our main contributions are threefold: 

\begin{itemize}
    \item \sloppy We introduce \textbf{RefChartQA}, a new dataset of real-world charts enriched with bounding-box annotations for answer grounding. It extends existing ChartQA resources \cite{masry-etal-2022-chartqa,zhang2024tinychart} by focusing on questions involve arithmetic or logical reasoning.
    \item We propose an \textbf{instruction-tuning} strategy that adapts multimodal large language models (LLMs) to simultaneously handle question-answering and bounding-box generation with minimal architectural overhead.
    \item We conduct extensive \textbf{experiments} demonstrating that our grounded approach significantly improves interpretability and can also help reduce hallucinations when dealing with complex numerical queries.
\end{itemize}

%% file: chapters/relatedwork.tex
\section{Related Work}
\label{sec:related_work}

\subsection{Chart QA Benchmarks}
\label{subsec:chart_qa_benchmarks}
A variety of Chart QA benchmarks have been proposed to evaluate a model’s ability to analyze and reason over data visualizations. Initially, researchers relied on \textit{synthetic} datasets like FigureQA \cite{kahou2017figureqa}, DVQA \cite{kafle2018dvqa} and \cite{li2024synthesize}, which use template-generated charts and questions. These studies demonstrated the feasibility of automating chart reasoning but often lacked the stylistic diversity and complexity of real-world charts. To address this, PlotQA \cite{methani2020plotqa} introduced open-vocabulary queries, capturing more advanced arithmetic or logical operations, yet remained inherently synthetic. Moving beyond templates, ChartQA \cite{masry-etal-2022-chartqa} presented over 9,600 human-curated questions covering diverse chart types and domains. Subsequent endeavors have explored automated generation via LLMs, e.g., ChartX \cite{xia2024chartx} and ChartLlama \cite{han2023chartllama}, or self-training methods, e.g., EvoChart \cite{huang2024evochart} to broaden variety and complexity. Despite these steps, explicit \emph{visual grounding} in chart contexts remains underexplored: only PlotQA includes limited bounding-box labels for basic tasks, highlighting a persistent need for benchmarks incorporating fine-grained answer localization.

\begin{figure}[t!]
    \centering
    \includegraphics[width=\linewidth]{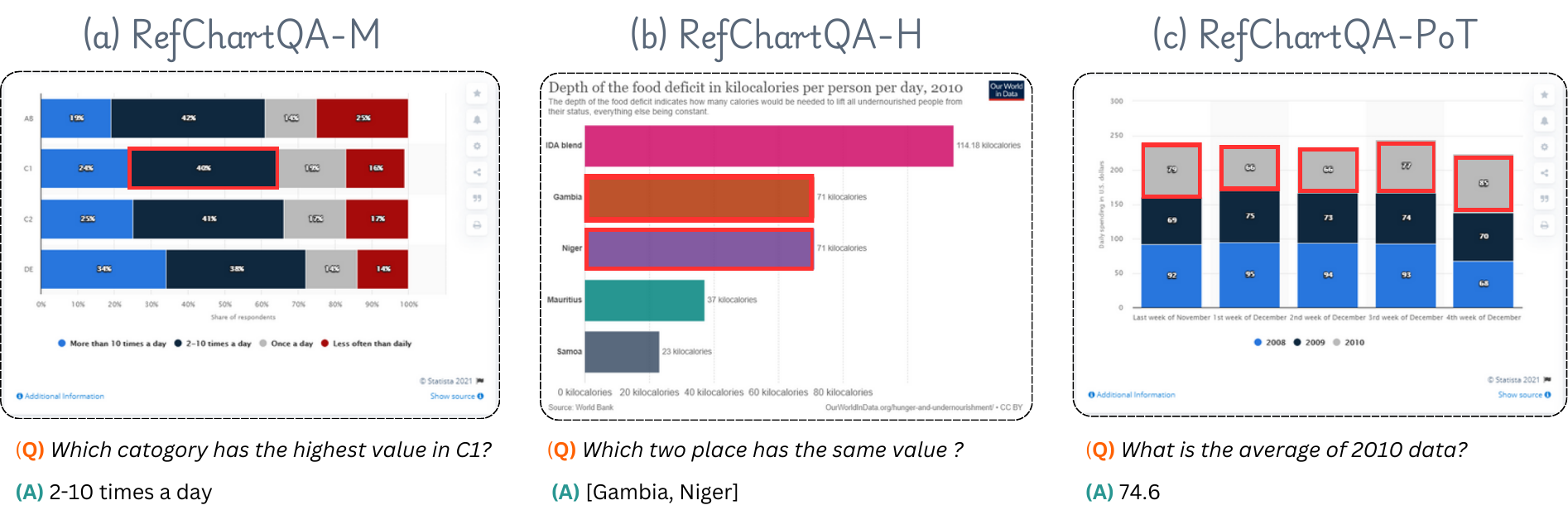}
    \caption{Examples from the RefChartQA benchmark: (a) A machine-generated question-answer pair, (b) A human-posed question, and (c) A PoT sample that requires mathematical reasoning and grounding of multiple elements.}
    \label{fig:samples}
\end{figure}

\subsection{Chart Understanding Models}
\label{subsec:chart_understanding_models}
Chart understanding has traditionally involved \textit{multi-stage pipelines}: extracting textual or numerical content via OCR and component detectors, then converting that content into tabular form for downstream QA \cite{methani2020plotqa,masry-etal-2022-chartqa, liu2023deplot}. While these methods yield interpretable intermediate results, the reliance on OCR can introduce errors, and the final QA often overlooks intricate visual details (colors, positions, etc.). More recently, \textit{end-to-end vision-language} models have gained traction for chart tasks. Examples include UniChart \cite{masry-etal-2023-unichart}, which adapts Donut \cite{kim2022donut} for chart comprehension, and Pix2Struct \cite{lee2023pix2struct}, which encodes visual features of chart screenshots for direct QA. State-of-the-art systems now harness \textit{instruction-tuned} multimodal large language models like LLaVA \cite{liu2023llava}, PaliGemma \cite{beyer2024paligemma, steiner2024paligemma2}, and ChartLlama \cite{han2023chartllama} to process complex queries. Additionally, specialized techniques address numerical reasoning by coupling code execution and large language models (Program-of-Thought), as demonstrated by TinyChart \cite{zhang2024tinychart} and ChartGemma \cite{masry-etal-2025-chartgemma}. These approaches significantly boost answer accuracy but typically provide limited transparency about \emph{which} regions of the chart were utilized for reasoning. Integrating explicit bounding-box-based \emph{visual grounding} remains a gap, motivating our work to unify high-level chart QA performance with localized, interpretable evidence.

\subsection{Visual Answer Grounding} 
\label{subsec:vg}
Visual Answer Grounding extends VQA by explicitly localizing image regions that justify predicted answers \cite{chen2023vqatherapy,khoshsirat2023sentence,lin2024lcv2}. While foundational studies on natural images employed region-proposal networks and attention-based mechanisms \cite{zhu2016visual7w,lu2019vilbert}, recent research emphasizes integrating large language models (LLMs) with external grounding modules \cite{ma2024groma,zhang2023llavagrounding,zhao2023bubogpt}. These architectures often target object referral or open-set detection and excel at visually explaining answers in photographs. However, when applied to chart images, where textual labels, numerical axes, and structured graphical elements play a pivotal role, most general-purpose VAG systems lack specialized handling of numerical reasoning and chart-specific semantics.

%% file: chapters/benchmark.tex
\section{RefChartQA Benchmark}
\begin{figure}[t!]
    \centering
    \includegraphics[width=\linewidth]{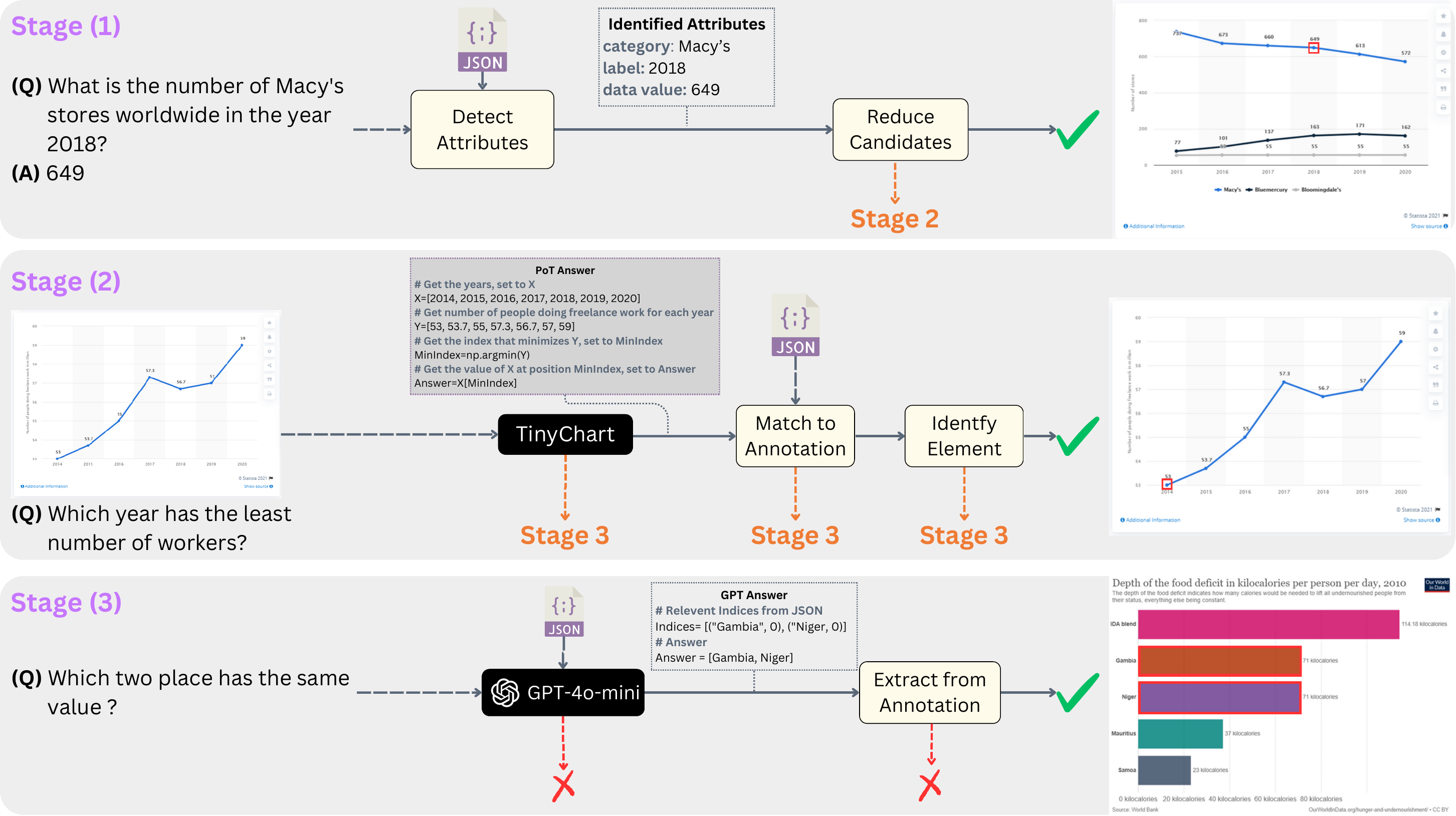}
    \caption{Overview of the stages in the RefChartQA annotation process. A sample is considered successfully annotated by a stage when all its steps are completed. Orange and red downward arrows indicate points where validation failures may lead to sample rejection at that stage.}
    \label{fig:annoprocess}
\end{figure}

To enable visual grounding for chart images, we introduce RefChartQA, a benchmark that extends ChartQA \cite{masry-etal-2022-chartqa} and a subset of TinyChart's ChartQA-PoT \cite{zhang2024tinychart} with explicit grounding annotations. Unlike prior datasets that focus solely on answer accuracy, RefChartQA ensures models justify their predictions by linking answers to relevant chart elements. Grounding annotations are generated through a multi-stage process, including a heuristic-based extraction, a Program-of-Thought (PoT)-based method, and a GPT-based method. The whole process is summarized in Fig. \ref{fig:annoprocess}. Next, we discuss how we formed the annotations and present the statistical analysis of our benchmark.

\subsection{Annotation Process}
We implement visual grounding using a bounding box paradigm, which is particularly effective for chart images due to its versatility across different chart types. However, we do not claim this to be the best or only approach; alternative methods, such as semantic segmentation or keypoint detection, could provide complementary advantages by capturing finer details or structural relationships within charts. These methods, however, often come with higher computational costs. For instance, autoregressive segmentation typically requires significantly more processing power and training time. While the keypoint paradigm is a more native approach, it is less suited for many common visualizations where answers correspond to regions rather than single points. Future work could explore multi-grounding paradigms that adapt based on the image and query context. In identifying grounding, we prioritize relevant data values, as they often implicitly indicate the associated structural information. A data element is included in the grounding if it is directly referenced in the answer or is essential for computing it. For min/max questions, we include only the identified element, while for operations like sum or average, all contributing elements are included. When the answer depends on structural aspects rather than specific data values, we annotate the relevant structural element.

We choose ChartQA \cite{masry-etal-2022-chartqa} as the foundation for our dataset due to its diverse question types, covering both arithmetic and non-arithmetic reasoning over chart images. Arithmetic questions often require multi-element operations, such as calculating averages or making comparisons, while non-arithmetic questions typically refer to a single chart element, though some may involve multiple elements for contextual understanding. ChartQA provides comprehensive structured metadata for each chart, including bounding boxes and textual content for both descriptive elements (such as axis labels and legend entries) and data visualization elements within the chart. Data elements are categorized and include color information. Additionally, we incorporate a subset of the ChartQA-PoT dataset \cite{zhang2024tinychart}, as it provides additional arithmetic questions based on ChartQA charts and structured reasoning in the form of Python code. Its use of ChartQA charts allows the reuse of ChartQA's metadata without modifying the annotation process. However, both datasets lack explicit connections between visual elements and the questions, as they do not include direct grounding annotations. To bridge this gap, we develop a structured annotation process that links questions to their supporting visual evidence as summarized in Fig. \ref{fig:annoprocess}.

The annotation pipeline is designed to match questions with relevant evidence elements from the metadata. As a result, our dataset may carry over errors from the underlying source datasets, as well as introduce new mistakes during the grounding annotation process. To minimize errors, we incorporate validation in the annotation process. 

\vspace{5pt}
\noindent \circled{1} \textbf{Single-Element Extraction.} The first stage of the annotation pipeline targets single-element non-arithmetic questions, where the answer corresponds directly to a single data element in the chart. Since these questions do not involve computational reasoning or the interaction of multiple elements, a heuristic-based approach is used to match answers with their corresponding visual elements. For example, the question \textit{"What is the number of Macy’s stores worldwide in the year 2018?"} describes a data retrieval operation which can be answered using structural information only. To identify the relevant evidence element, the question and answer text are analyzed to extract x-axis labels, categories, and data values (from the answer only). String- and color matching, based on metadata, are used to identify the referenced elements. The extracted attributes are used to constrain the set of data element candidates. If a single match remains, it is selected as the grounding element. Otherwise, the annotation fails and the question progresses to the next stage. 

\vspace{5pt}
\noindent \circled{2} \textbf{PoT Extraction.} The second stage introduces PoT-based reasoning to handle more complex question types, particularly those requiring arithmetic operations or multi-element interactions. Unlike the first stage, which relies on direct string-matching, this stage leverages TinyChart’s PoT \cite{zhang2024tinychart} to reconstruct the logical reasoning path necessary to derive the grounding. For example, consider the following PoT code: 

\begin{listing}[h]
\centering
\vspace{-3ex}
\label{lst:pseudocode_flow_with_comments}
\begin{mdframed}[
    linewidth=1pt,
    roundcorner=5pt,             
    backgroundcolor=white,   
    linecolor=black
]
\textbf{Question:}~\textit{"Which year has the least number of workers?"}
\begin{minted}{python}
# Get the years, set to X
X=[2014, 2015, 2016, 2017, 2018, 2019, 2020]
# Get number of people doing freelance work for each year
Y=[53, 53.7, 55, 57.3, 56.7, 57, 59]
# Get the index that minimizes Y, set to MinIndex
MinIndex=np.argmin(Y)
# Get the value of X at position MinIndex, set to Answer
Answer=X[MinIndex]
\end{minted}
\end{mdframed}
\vspace{-5ex}
\end{listing}

The PoT process above starts by extracting the relevant metadata needed to answer the question—specifically, the list of years \texttt{X} and the number of people doing freelance work each year \texttt{Y}. This is followed by a sequence of reasoning operations, culminating in the final computed answer, which is stored in the variable \texttt{Answer}.
To ensure consistency with ChartQA annotations, we first align the extracted list values with the chart's metadata. This step establishes a mapping between approximated values and metadata values while also verifying the accuracy of the extracted elements. To further minimize annotation errors, we enforce that the PoT prediction must exactly match the ground truth answer.
After validation, we extract the relevant reasoning elements using a template-based approach. The corresponding metadata elements serve as grounding annotations. If validation fails or the reasoning complexity exceeds the capabilities of the template-based approach, the question advances to the next stage.

\vspace{5pt}
\noindent \circled{3} \textbf{GPT-based Annotation.}
To annotate more complex reasoning questions, where the single-element and PoT-based annotation methods fail, we use a large language model. With the chart metadata we are able to transform the relevant element extraction problem from a multimodal to a text-only problem. We prompt the GPT-4o-mini model with the question and the chart’s metadata, which includes the data values and visual attributes. The model is then tasked with predicting the answer and identifying the key chart metadata elements essential to the reasoning. Similar to the PoT approach, we again require exact correctness of the prediction to reduce hallucination. 
The goal of this stage is to extend the dataset to include more complex reasoning questions. In the following section, we analyze the statistical characteristics of our benchmark.

\newpage
\subsection{Data Statistics}
Since our benchmark is built upon the existing ChartQA dataset, which consists of human-authored (H) and machine-generated (M) samples, we preserve the original data splits. ChartQA-PoT was originally designed as a training dataset; therefore, we apply the 70/10/20 principle to partition the annotated samples into training, validation, and test sets. The resulting combined RefChartQA consists of 73,702 image-question-grounding pairs, with 55,789 for training, 6,223 for validation, and 11,690 for testing. Fig. \ref{fig:samples} shows samples from each of the benchmark splits. The distribution of annotation success across different stages: single-element retrieval, PoT-based reasoning, and GPT-based grounding, as presented in Table \ref{tab:annotation_success}. 

To evaluate annotation quality, we conducted manual validation on a subset of the dataset. 50 samples from each split were randomly selected and reviewed by two independent annotators, resulting in a 93.45\% agreement rate on grounding correctness. However, certain cases, particularly in ChartQA-H, had multiple valid grounding solutions. For example, in stacked bar charts with questions like \textit{"Which category contributes the most to the total?"}, one annotator might highlight the legend element, while another might select the entire bar as a valid reference. These findings underscore the inherent challenges of document grounding and motivate us to present this benchmark to further encourage research in this area.

\begin{table}[t]
    \centering
    \caption{Overview of RefChartQA annotation success rates across the three stages. The \textbf{Total} column is highlighted to emphasize the final dataset size and success percentage.}
    \label{tab:annotation_success}
    \renewcommand{\arraystretch}{1.2}
    \setlength{\tabcolsep}{6pt}  
    \newcolumntype{T}{>{\columncolor[gray]{0.95}}c}  
    \resizebox{\linewidth}{!}{   
    \begin{tabular}{lccccT}  
        \toprule
        \textbf{Dataset} & \multicolumn{1}{c}{\textbf{\# Questions} $\rightarrow$} & \multicolumn{1}{c}{\textbf{Single Element} $\rightarrow$} & \multicolumn{1}{c}{\textbf{PoT-based} $\rightarrow$} & \multicolumn{1}{c}{\textbf{GPT-based}} & \multicolumn{1}{c}{\textbf{Total (\%)}} \\
        \midrule
        \textit{ChartQA-M}  & 23,111  & 16,071  & 3,032  & 0  & \textbf{19,103 (82.7\%)}  \\
        \textit{ChartQA-H}  & 9,608   & 1,430   & 1,616  & 767 & \textbf{3,813 (39.7\%)}  \\
        \textit{ChartQA-PoT} & 119,281  & 24,181  & 10,864 & 15,741 & \textbf{50,786 (42.6\%)}  \\
        \bottomrule
    \end{tabular}}

\end{table}

%% file: chapters/method.tex
\section{Methodology}
\label{sec:method}

Building on the pixel-to-sequence paradigm, we implement VAG using transformers to encode visual input as a structured sequence of tokens for efficient processing. Similarly to Pix2Seq \cite{chen2021pix2seq}, which reformulates object detection using a transformer-based autoregressive decoder, this approach enables flexible sequence generation for spatially structured data. Subsequently, Pix2Seq (v2) \cite{chen2022unified} extends this method by introducing instruction-based tuning, enabling a unified approach to multiple vision tasks without requiring task-specific heads and objectives. Chart-specific models such as ChartGemma \cite{masry-etal-2025-chartgemma}, ChartLlama \cite{han2023chartllama}, and TinyChart \cite{zhang2024tinychart} follow a similar instruction-tuned multitask framework, where task-specific prompts shape language outputs. Building on this foundation, we experiment with multimodal large language models to integrate spatial awareness while preserving core interpretive skills.

\subsection{Grounding Sequence Construction}
We implement visual grounding using a bounding box paradigm. To achieve this, we leverage the existing language tokens to encode the bounding boxes in a corner-based representation, enclosing each bounding box with $\text{<box>}\dots\text{</box>}$. This approach enables the LLMs to adapt to the new task while maintaining their original task-specific skills. Furthermore, interleaving spatial and textual information in autoregressive models has been shown to improve image understanding by improving the model’s ability to associate objects with their contextual descriptions \cite{lu2024bounding}. To prevent inconsistencies in spatial representation, we follow Qwen-VL \cite{wang2024qwen2} approach and normalize and quantize each coordinate to [0, 1000).

Unlike natural images, which often lack inherent layout, visualizations follow common spatial structures, with data elements grouped into categories and arranged in meaningful sequences, such as left-to-right (e.g., x-axis labels) or top-down (e.g., stacked bar segments and hierarchical labels in tree maps). In autoregressive models, the sequence in which bounding box parameters are predicted can significantly impact the model's performance \cite{liu2022autoregressive}, as each prediction is conditioned on the previously predicted components. While models like Pix2Seq assume random ordering for object detection tasks, our grounding annotation preserves this intrinsic order. Our final output template below is structured to accommodate multiple $n$ bounding boxes, with an ad-hoc separator to distinguish between the grounding information and the final answer.

\begin{center}
\makebox[\textwidth]{\textit{(<box> $x_{\text{min}}, y_{\text{min}}, x_{\text{max}}, y_{\text{max}}$ </box>)$^n$ | <\text{grounding-sep}> | \text{answer}}}
\end{center}

%% file: chapters/experiment.tex
\section{Experiment}
\label{sec:experiments}
We evaluate our benchmark through instruction tuning using several large vision-language models from different categories. Section \ref{model_selection} begins by introducing our model selection. Section \ref{exp_setup} introduces the experimental setup, followed by a discussion of the evaluation metrics in Section \ref{metrics}. Next, we report the performance comparison through comprehensive quantitative evaluations in Section \ref{quan}. Finally, we analyze case studies and provide a qualitative discussion of our observations in Section \ref{qual}.

\subsection{Model Selection}
\label{model_selection}
To evaluate the impact of different pretrained models, we selected several SOTA MLLMs across chart analysis, document understanding, and visual grounding. For chart-related models, we chose UniChart \cite{masry-etal-2023-unichart}, ChartGemma \cite{masry-etal-2025-chartgemma}, and TinyChart \cite{zhang2024tinychart}. UniChart, a lightweight Donut-based \cite{kim2022donut} model with 260M parameters, delivers competitive performance despite its smaller size. In contrast, ChartGemma and TinyChart ($\sim$3B parameters) incorporate advanced reasoning capabilities. TinyChart supports two inference modes: direct inference for end-to-end predictions and PoT-based reasoning, which generates executable Python code. ChartGemma, however, exclusively relies on PoT-based predictions. All three models have been instruction-tuned on chart-related tasks like summarization, chart-to-table conversion, and chart VQA.
For visual grounding, we selected Qwen2.5-VL \cite{Qwen2.5-VL} and Qwen-VL-Chat \cite{wang2024qwen2}, one of the few VG models fine-tuned on ChartQA. Alongside image captioning, VQA, and OCR, it has been trained on grounded image captioning and object referral, making it a strong candidate for grounding-based evaluations.
For document understanding, we included DocOwl 1.5 \cite{hu2024docowl}, which, like chart-focused models, has been trained on document parsing, table extraction, chart parsing, and VQA, while also incorporating text localization.

\input{tables/main_table}
\subsection{Experiment Setup}
\label{exp_setup}
We conducted our experiments on four Nvidia A40 GPUs with 48 GB of VRAM. All models were fine-tuned by parameter-efficient fine-tuning using LoRA adapters \cite{hu2022lora} for five epochs with a global batch size of 16. For UniChart, we used full-parameter instead of LoRA tuning due to its comparatively small parameter count. For the full configuration details, please refer to Table~\ref{tab:training_params}. At this point, we treat the bounding box predictions as text and use the next-token prediction objective, with a softmax cross-entropy loss.
For inference we relied on direct inference mode and the following input prompt (with slight variances depending on how the model processes images):

\begin{center}
  \itshape
  “\textit{}{<image>\{question\}  Append grounding bounding boxes.}”
\end{center}

\noindent For UniChart, a special ground-token task was introduced:

\begin{center}
  \itshape
  “\textit{<grounding\_chartqa>\{question\} <s\_answer>}”
\end{center}

\input{tables/config_table}

\subsection{Evaluation Metrics}
\label{metrics}

\noindent\textbf{QA Metric.} To assess the response correctness, we adopt the \textbf{Relaxed Accuracy}~\cite{masry-etal-2022-chartqa}, commonly used in Chart-QA tasks. This metric allows a maximal $5\%$ error margin for floating-point numeric predictions, while an exact match is required for string answers.  

\noindent\textbf{Grounding Metric.}
Average Precision, AP@0.5, is a common metric reported for object detection evaluation. It is also suitable for chart grounding since chart elements rarely overlap, and an IoU $\geq$ 0.5 is generally sufficient to assign predictions to the correct chart element. However, AP@0.5 alone is not sufficient. The metric aggregates precision and recall across the entire dataset rather than evaluating correctness on a per-sample basis. For example, a model might achieve high AP@0.5 by correctly grounding many easy cases while still failing on harder multi-element grounding tasks. 
To address this, we report Precision@F$_1$=1, IoU$\geq$0.5 (P@FI) \cite{he2023grec}, a metric originally introduced for Generalized Referring Expression Comprehension (GREC). P@FI ensures that all groundings for a given question must be true for the sample to be counted as correct.

\subsection{Quantitative Analysis}
\label{quan}
\noindent\textbf{Finding \circled{1} : MLLMs exhibit strong visual-attention grounding but still struggle with complex queries.} 
Table \ref{tab:model_compare} demonstrates that models of different expertise achieve reliable grounding performance, which emerges after a few training epochs. The models attain P@FI $\approx$ 70–84\% in RefChartQA-M, which consists mainly of simple single element queries. However, performance declines in RefChartQA-PoT (P@FI $\approx$ 40–70\%), where queries require multi-element grounding and more intricate reasoning. The performance further decreases in RefChartQA-H (P@FI $\approx$ 22–50\%), which consists of human-posed questions that more closely reflect real-world applications. These results suggest that while foundational grounding abilities emerge early during training, more sophisticated tuning is required to address real-world challenges involving reasoning-grounding interdependencies.

\vspace{4pt}
\noindent\textbf{Finding \circled{2}: MLLMs could achieve fewer hallucinations with grounding.} 
Models across various categories generally improve response accuracy when grounding is incorporated, particularly in the more complex human-posed and PoT splits. Notably, TinyChart, Qwen-VL-Chat, and DocOwl achieve up to a +15\% increase in relaxed accuracy with the grounding task. The reason behind this significant improvement lies in a common pitfall observed in model behavior—as shown in Fig. \ref{fig:casestudy}-(a), models often engage in unnecessary comprehensive reasoning for queries that only require direct data retrieval. Hence, grounding may help mitigate the need for such redundant reasoning loops, and effectively encoding spatial information in the model’s embeddings could further reduce hallucinations.

\vspace{4pt}
\noindent\textbf{Finding \circled{3}: Larger models do not always guarantee better grounding performance.}
While the document-expert model (DocOwl 1.5-8B) and the grounding-expert model (Qwen-VL-Chat 9.6B) contain significantly more parameters and a larger pretraining corpus than TinyChart-3B, their grounding performance does not consistently surpass that of the smaller model. This highlights that effective grounding depends on how well the model inherits spatial information from the source domain. We believe that model architecture plays a more crucial role than the pretraining corpus, as domain-specific expert models such as UniChart and ChartGemma also exhibit a significant decline in $acc_g$ with grounding. For example, in the case of TinyChart, it adopts a parameter-free Visual Token Merging module \cite{bolya2022token} within each vision transformer layer, which aggregates the most similar visual tokens—potentially contributing to spatial knowledge formation. However, this observation requires rigorous experimentation to be conclusively validated.


\subsection{Qualitative Analysis}
\label{qual}
In addition to the quantitative performance analysis, we aimed to qualitatively examine scenarios where introducing VAG task outperformed the model w/o grounding and to explore the correlation between text response and detection performance.

\noindent\textbf{Case Study \circled{1} : Correct Answer with Grounding.} One of the key observations in our analysis is the tendency of models, particularly those requiring PoT reasoning, to generate unnecessarily complex thought-chains in the absence of grounding. As illustrated in Figure \ref{fig:casestudy}-(a-b), these models often struggle to determine whether they should perform complex calculations or directly extract the answer from the chart image. Introducing grounding enhanced the model's interpretative capabilities by incorporating spatial information. This improvement can be likened to a finger pointing toward the relevant answer within the chart, directing the model’s attention to the critical regions of the image.

\vspace{4pt}
\noindent\textbf{Case Study \circled{2}: Correct Grounding But Wrong Answer.} In several observed cases, the model was able to accurately localize relevant information within the chart but failed to combine and reason to form the correct answer. One of the most common reasons for this failure is a model’s deficiency in mathematical reasoning. For example, the model identifies multiple relevant elements within the chart but struggles to perform a sum operation over these elements as in Fig. \ref{fig:casestudy}-(c-d).

\noindent\textbf{Case Study \circled{3}: Incorrect Grounding and Answer.} 
Another common pattern observed is when both the predicted answer and the bounding box are incorrect. We identify three key reasons for these errors:
\begin{itemize}
    \item In some cases, Fig. \ref{fig:casestudy}-(e) the predicted answer and grounding were internally consistent and aligned but failed to address the actual question scope, leading to an irrelevant response.
    
    \item Some errors arise due to the possibility of multiple valid answers for a given question, whereas our annotation process favors a single correct option as in Fig. \ref{fig:casestudy}-(f).
    
    \item Although less frequent, a few instances were observed where both the predicted answer and bounding box were entirely nonsensical, highlighting a reasoning-grounding alignment failure.
    
\end{itemize}
\input{chapters/casestudy_figure}

%% file: tables/main_table.tex
\definecolor{mygreen}{RGB}{0,150,0} 
\definecolor{myred}{RGB}{200,0,0} 

\begin{table}[!t]
    \centering
    
    \caption{Comparison of several MLLMs on the RefChartQA benchmark across three test splits: human, machine, and PoT. Relaxed accuracies are reported both w/o grounding finetuning (acc$_\text{o}$) and with grounding (acc$_\text{g}$). Response accuracy gain is highlighted in \textcolor{mygreen}{green}, while degradations are highlighted in \textcolor{myred}{red}.}

    \label{tab:model_compare}
    \renewcommand{\arraystretch}{1.3} 
    \setlength{\tabcolsep}{1pt} 
    \resizebox{\textwidth}{!}{
    \begin{tabular}{l|c|c|>{\centering\arraybackslash}p{1cm}ccc|>{\centering\arraybackslash}p{1cm}ccc|>{\centering\arraybackslash}p{1cm}ccc}
        \hline
        \multicolumn{3}{c}{} & \multicolumn{4}{|c|}{\textbf{RefChartQA-H}} & \multicolumn{4}{c|}{\textbf{RefChartQA-M}} & \multicolumn{4}{c}{\textbf{RefChartQA-PoT}}\\
        \hline
        Model & \#Param. & Resolution & acc$_\text{o}$ & acc$_\text{g}$ & AP@$0.5$ & P@FI & acc$_\text{o}$ & acc$_\text{g}$ & AP@$0.5$ & P@FI & acc$_\text{o}$ & acc$_\text{g}$ & AP@$0.5$ & P@FI \\
        \hline
        
        \rowcolor[HTML]{F7F7F7} \multicolumn{15}{l}{\textbf{\textit{Chart-related Models}}} \\
        UniChart \cite{masry-etal-2023-unichart} & 260M & - & 58.27 & 51.40 \textcolor{myred}{(-06.87)} & 18.30 & 31.60 & 92.93 & 88.47 \textcolor{myred}{(-04.46)} & 57.39 & 75.58 & 72.17 & 64.49  \textcolor{myred}{(-07.68)} & 44.68 & 53.37 \\
        ChartGemma \cite{masry-etal-2025-chartgemma} & 2B & 448$\times$448 & 82.48 & 71.20 \textcolor{myred}{(-11.28)} & 19.95 & 39.00 & 93.99 & 94.67 \textcolor{mygreen}{(+00.68)} & 60.62 & 77.91 & 80.43 & 72.05 \textcolor{myred}{(-08.38)} & 43.44 & 54.92 \\
        TinyChart-PoT \cite{zhang2024tinychart} & 3B & 768$\times$768 & 86.40 & \textbf{86.02} \textcolor{myred}{(-00.38)} & 25.47 & 46.65 & 97.01 & 96.22 \textcolor{myred}{(-00.79)} & 66.10 & 81.88 & \textbf{93.03} & \textbf{95.40} \textcolor{mygreen}{(+02.37)} & 58.27 & 66.63 \\
        TinyChart \cite{zhang2024tinychart} & 3B & 768$\times$768 & 58.96 & 76.38 \textcolor{mygreen}{(+17.42)} & 27.81 & 49.41 & 94.48 & \textbf{96.61} \textcolor{mygreen}{(+02.13)} & \textbf{71.25} & \textbf{84.30} & 66.90 & 82.26 \textcolor{mygreen}{(+15.36)} & \textbf{59.66} & \textbf{67.58} \\ 
        \hline
        
        \rowcolor[HTML]{EFEFEF} \multicolumn{15}{l}{\textbf{\textit{Grounding Models}}} \\
        Qwen-VL-Chat \cite{wang2024qwen2} & 9.6B & 448$\times$448 & 50.98 & 54.33 \textcolor{mygreen}{(+03.35)} & 27.51 & 44.69 & 79.26 & 89.53 \textcolor{mygreen}{(+10.27)} & 64.81 & 80.72 & 40.33 & 63.27 \textcolor{mygreen}{(+22.94)} & 47.53 & 57.03 \\
    Qwen2.5-VL \cite{Qwen2.5-VL} & 3B & native res. & \textbf{88.80} & 84.80 \textcolor{myred}{(-04.00)} & \textbf{32.83} & \textbf{50.40} & \textbf{97.30} & 96.60 \textcolor{myred}{(-00.70)} & 59.28 & 76.12 & 80.60 & 81.73 \textcolor{mygreen}{(+01.13)} & 39.32 & 46.86 \\

        \hline
        
        \rowcolor[HTML]{F7F7F7} \multicolumn{15}{l}{\textbf{\textit{Document Models}}} \\
        DocOwl 1.5 \cite{hu2024docowl} & 8B & 448$\times$448($\times$9) & 56.30 & 59.84 \textcolor{mygreen}{(+03.54)} & 18.53 & 35.63 & 95.06 & 94.28 \textcolor{myred}{(-00.78)} & 54.08 & 73.64 & 49.05 & 68.39 \textcolor{mygreen}{(+19.34)} & 30.86 & 45.37 \\
        \hline
    \end{tabular}
    }
\end{table}

%% file: tables/config_table.tex
\begin{table}[ht!]
  \centering
  \caption{Training configurations for all models.}
  \label{tab:training_params}
  \renewcommand{\arraystretch}{1.2}   
  \setlength{\tabcolsep}{3pt}        
  \footnotesize                       
  \resizebox{0.95\textwidth}{!}{%
    \begin{tabular}{l c c c c c c c c c}
      \hline
      \textbf{Model}    & \textbf{Approach}    & \textbf{Epochs} & \textbf{Batch} & \textbf{LR}    & \textbf{Optimizer} & \textbf{Scheduler} & \(\mathbf{r}\) & \(\boldsymbol\alpha\) & \textbf{Dropout} \\
      \hline
      UniChart           & Full param.        & 5               & 16             & 5.00E-05      & Adam               & cosine             & –              & –                     & –               \\
      ChartGemma         & LoRA               & 5               & 16             & 1.00E-04      & AdamW              & –                  & 8              & 8                     & 0               \\
      TinyChart-PoT      & LoRA               & 5               & 16             & 1.00E-04      & AdamW              & cosine             & 64             & 16                    & 0.05            \\
      TinyChart          & LoRA               & 5               & 16             & 1.00E-04      & AdamW              & cosine             & 64             & 16                    & 0.05            \\
      QwenVL-Chat        & LoRA               & 5               & 16             & 1.00E-05      & AdamW              & cosine             & 64             & 16                    & 0.05            \\
      QwenVL-2.5         & LoRA               & 1               & 16             & 2.00E-04             & AdamW                  & cosine                  & 16              & 16                     & 0.05               \\
      DocOwl1.5          & LoRA               & 5               & 16             & 2.00E-05      & AdamW              & cosine             & 128            & 256                   & 0.05 \\
      \hline
    \end{tabular}%
  }
\end{table}

%% file: chapters/casestudy_figure.tex
\begin{figure*}
\vspace{-2ex} 
\centering
\renewcommand{\arraystretch}{1.2} 
\setlength{\tabcolsep}{10pt}
\resizebox{0.85\textwidth}{!}{%
\begin{tabular}{M{0.48\textwidth} M{0.48\textwidth}}
    \begin{minipage}{\linewidth}
        \raggedright
        \includegraphics[width=\linewidth]{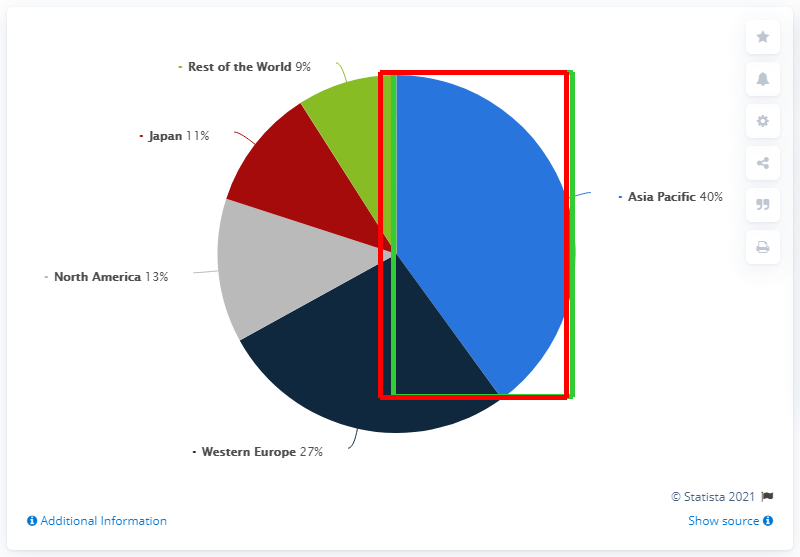}  
        \vspace{4pt}
        \textbf{Question:} \textit{What is the ratio of Asia Pacific in Revenue share in this pie chart?} \\
        \textbf{ground-truth :} 40 \\
        \textbf{TinyChart :} \textcolor{red}{2.047619048} \\
        \textbf{TinyChart\textsubscript{ground} :} \textcolor{mygreen}{40} \\
        \vspace{4pt}
        \centering \textit{(a)}
    \end{minipage}
    &
    \begin{minipage}{\linewidth}
        \raggedright
        \includegraphics[width=\linewidth]{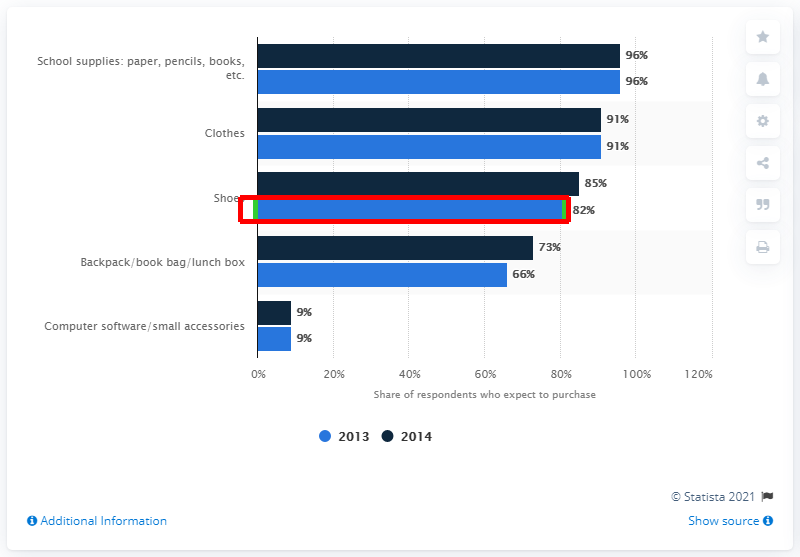}  
        \vspace{4pt}
        \textbf{Question:} \textit{What was the 3rd most popular item in 2013?} \\
        \textbf{ground-truth :} Shoes \\
        \textbf{TinyChart :} \textcolor{red}{Backpack/book/lunch-box} \\
        \textbf{TinyChart\textsubscript{ground} :} \textcolor{mygreen}{Shoes} \\
        \vspace{4pt}
        \centering \textit{(b)}
    \end{minipage} \\
    
    \begin{minipage}{\linewidth}
        \raggedright
        \includegraphics[width=\linewidth]{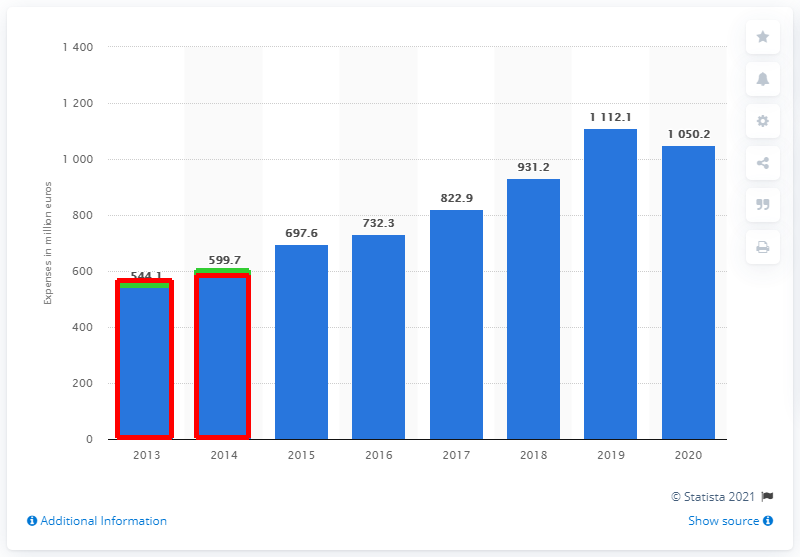}  
        \vspace{4pt}
        \textbf{Question:} \textit{What is the sum of the two smallest expenses in million euros?} \\
        \textbf{ground-truth :} 1143.8 \\
        \textbf{TinyChart :} \textcolor{red}{1213.7} \\
        \textbf{TinyChart\textsubscript{ground} :} \textcolor{orange}{1093.8} \\
        \vspace{4pt}
        \centering \textit{(c)}
    \end{minipage}
    &
    \begin{minipage}{\linewidth}
        \raggedright
        \includegraphics[width=\linewidth]{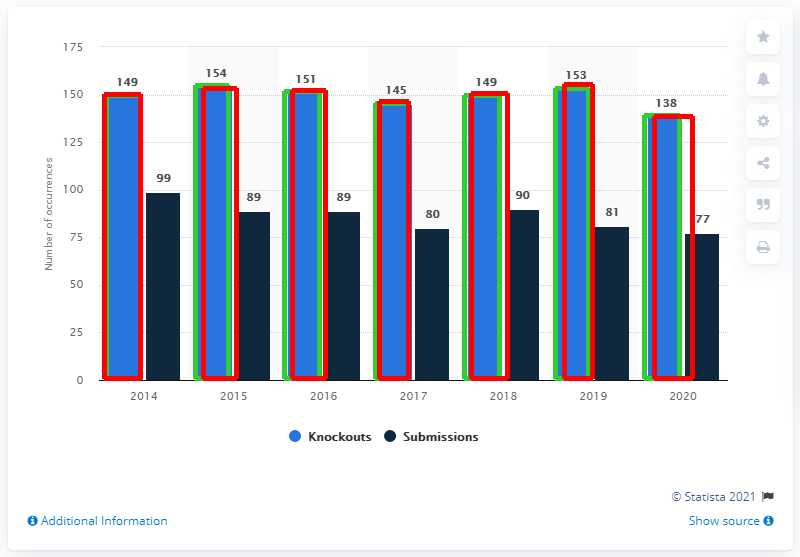}  
        \vspace{4pt}
        \textbf{Question:} \textit{What is the sum of Knockouts?} \\
        \textbf{ground-truth :} 1039 \\
        \textbf{TinyChart :} \textcolor{red}{905} \\
        \textbf{TinyChart\textsubscript{ground} :} \textcolor{orange}{1036} \\
        \vspace{4pt}
        \centering \textit{(d)}
    \end{minipage} \\
    
    \begin{minipage}{\linewidth}
        \raggedright
        \includegraphics[width=\linewidth]{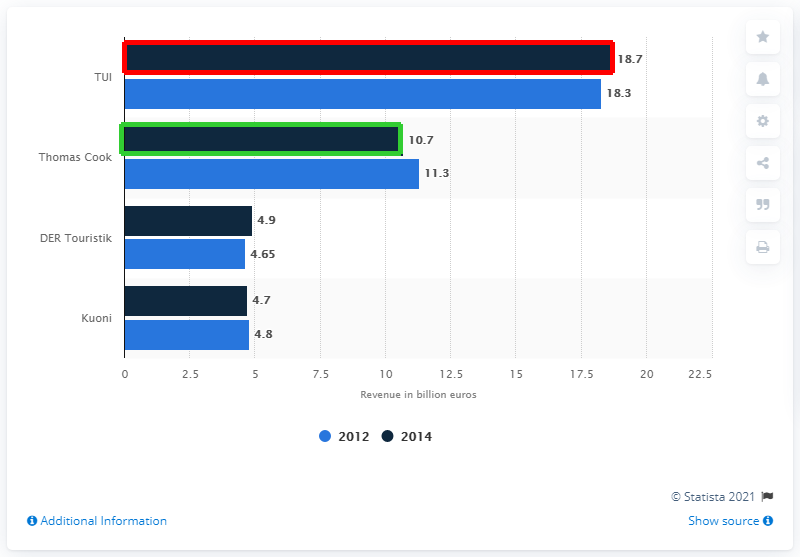}  
        \vspace{4pt}
        \textbf{Question:} \textit{How much did accounts with 1,000 to 10,000 followers increase their followers on average?} \\
        \textbf{ground-truth :} 12.6 \\
        \textbf{TinyChart\textsubscript{ground} :} \textcolor{red}{15.9} \\
        \vspace{4pt}
        \centering \textit{(e)}
    \end{minipage}
    &
    \begin{minipage}{\linewidth}
        \raggedright
        \includegraphics[width=\linewidth]{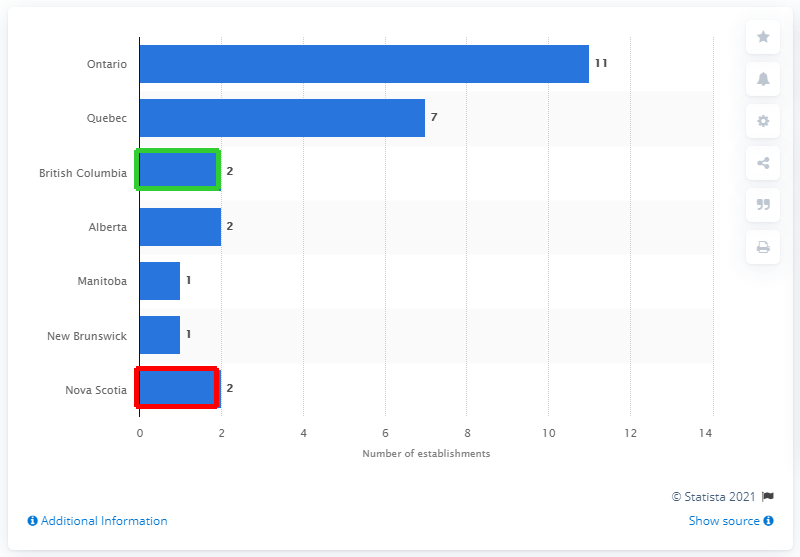}  
        \vspace{4pt}
        \textbf{Question:} \textit{In what province were there two roasted nut and peanut butter manufacturing establishments?} \\
        \textbf{ground-truth :} British Columbia \\
        \textbf{TinyChart\textsubscript{ground} :} \textcolor{purple}{Nova Scotia} \\
        \vspace{4pt}
        \centering \textit{(f)}
    \end{minipage} \\
\end{tabular}}
\vspace{4pt}
\caption{Qualitative case studies with TinyChart w/ and w/o VAG. 
\textcolor{myred}{Red} bounding boxes represent model predictions, 
while \textcolor{mygreen}{green} denote ground-truth annotations.}
\label{fig:casestudy}
\vspace{-2ex}
\end{figure*}

%% file: chapters/conclusion_and_future_work.tex
\section{Conclusion}
In this paper, we introduced \textit{RefChartQA}, the first benchmark that integrates chart question answering with visual answer grounding to enhance interpretability and response reliability in chart-based reasoning. Through comprehensive experiments on six state-of-the-art vision-language models across different categories, we demonstrate that incorporating grounding improves response accuracy, while also identifying key factors influencing text-spatial alignment. Despite these advancements, challenges remain in handling complex multi-element reasoning and numerical inference. We hope this work will drive further research in chart understanding and VAG, fostering more robust, interpretable, and spatially aware document understanding models.

\subsubsection{Limitations}
Despite its promising results, RefChartQA has limitations in data diversity and model generalization. While the benchmark incorporates various chart types, real-world visualizations often exhibit greater variability in design, annotation styles, and noise levels. Furthermore, the lack of standardized guidelines for grounding annotations among annotators has introduced some inconsistencies in the labels.
Finally, the role of architectural design in leveraging VAG remains underexplored. Large-scale VLMs differ in their ability to integrate spatial and textual cues, yet rigorous controlled experiments are needed to systematically analyze how different architectures benefit from grounding signals.

\subsubsection{Acknowledgment} This work was supported in part by Helmholtz Association of German Research Centers, in part by the Ministry of Science, Research and the Arts of Baden-Württemberg (MWK) through the Cooperative Graduate School Accessibility through AI-based Assistive Technology (KATE) under Grant BW6-03, and in part by Karlsruhe House of Young Scientists (KHYS). This work was partially performed on the HoreKa supercomputer funded by the MWK and by the Federal Ministry of Education and Research, partially on the HAICORE@KIT partition supported by the Helmholtz Association Initiative and Networking Fund, and partially on bwForCluster Helix supported by the state of Baden-Württemberg through bwHPC and the German Research Foundation (DFG) through grant INST 35/1597-1 FUGG.